\documentclass[letterpaper, 10 pt, conference]{ieeeconf}
\IEEEoverridecommandlockouts
\usepackage[bookmarks=true,hidelinks]{hyperref}
\usepackage[utf8]{inputenc}
\usepackage{color}
\usepackage{amsmath}
\usepackage{float}
\usepackage{graphicx}
\graphicspath{ {figures/} }

\usepackage{cite}
\usepackage{url}
\renewcommand{\UrlFont}{\small\tt}
\newcommand{\AtwoD}{\ensuremath{A_{n,2}\textsubscript{D}}}
\newcommand{\AthreeD}{\ensuremath{A_{n,3}\textsubscript{D}}}
\usepackage[utf8]{inputenc}
\usepackage[english]{babel}
\usepackage[dvipsnames]{xcolor}
\usepackage{balance}
\definecolor{myorange}{RGB}{195, 123, 71}
\usepackage{caption}
\title{\LARGE \bf Enumeration of Polyominoes \& Polycubes Composed of Magnetic Cubes%
\thanks{This work used
HPE DSI/IT computational resources and was
supported by the NSF under Grant Nos.\ \href{http://nsf.gov/awardsearch/showAward?AWD_ID=1553063}{[IIS-1553063,} 
\href{https://nsf.gov/awardsearch/showAward?AWD_ID=1849303}{IIS-1849303,}
\href{https://nsf.gov/awardsearch/showAward?AWD_ID=1932572}{CNS-1932572]}
.}%
\thanks{$^1$Authors are with the 
University of Houston, Houston, TX 77204 USA        {\tt\small  \{ylu36, debiediger, atbecker\}@uh.edu}.}%
\thanks{$^2$Authors are with  
Southern Methodist University, Dallas, TX 75275 USA        {\tt\small  \{abhattacharjee, mjkim\}@lyle.smu.edu}.}%
}
\author{Yitong Lu$^1$, Anuruddha Bhattacharjee$^2$, Daniel Biediger$^1$, Min Jun Kim$^2$, Aaron T. Becker$^1$}
\date{May 2020}
\begin{document}

\maketitle

\begin{abstract} 
This paper examines a family of designs for magnetic cubes and counts how many configurations are possible for each design as a function of the number of modules.
\emph{Magnetic modular cubes} are cubes with magnets arranged on their faces. The magnets are positioned so that each face has either magnetic south or north pole outward. Moreover, we require that the net magnetic moment of the cube passes through the center of opposing faces. These magnetic arrangements enable coupling when cube faces with opposite polarity are brought in close proximity and enable moving the cubes by controlling the orientation of a global magnetic field. This paper investigates the 2D and 3D shapes that can be constructed by magnetic modular cubes, and describes all possible magnet arrangements that obey these rules. We select ten magnetic arrangements and assign a ``color'' to each of them for ease of visualization and reference. 
We provide a method to enumerate the number of unique polyominoes and polycubes that can be constructed from a given set of colored cubes. 
We use this method to enumerate all arrangements for up to 20 modules in 2D and 16 modules in 3D.
We provide a motion planner for 2D assembly and through simulations compare which arrangements require fewer movements to generate and which arrangements are more common.
Hardware demonstrations explore the self-assembly and disassembly of these modules in 2D and 3D.
\end{abstract}

\section{Introduction}

Small-scale modular robots are often designed to be manipulated in large groups by an external system. Magnetic forces are a popular choice for manipulating these robots because the robots can be directly controlled simply by incorporating magnets or ferromagnetic material in the robot body. In the last few decades, many research projects have been conducted based on magnetic actuation and coupling systems to explore reconfigurable modular robotics at different length scales~\cite{romanishin2013m,tasoglu2014guided,xie2019reconfigurable,saab2019review}. Researchers have developed modules with excellent locomotion and manipulation abilities, and have integrated advanced control algorithms. Several reconfiguration strategies of modules, such as autonomous and distributed stochastic self-assembly, disassembly, self-reconfiguration by rotation in a planar workspace, locomotion through reconfiguration on a cubic lattice, and universal and distributed reconfiguration planning for square and hexagonal-lattice-based robots have been investigated~\cite{pfeifer2007self,tomita1999self}.

\begin{figure}[t]
\centering
\includegraphics[width=1\linewidth]{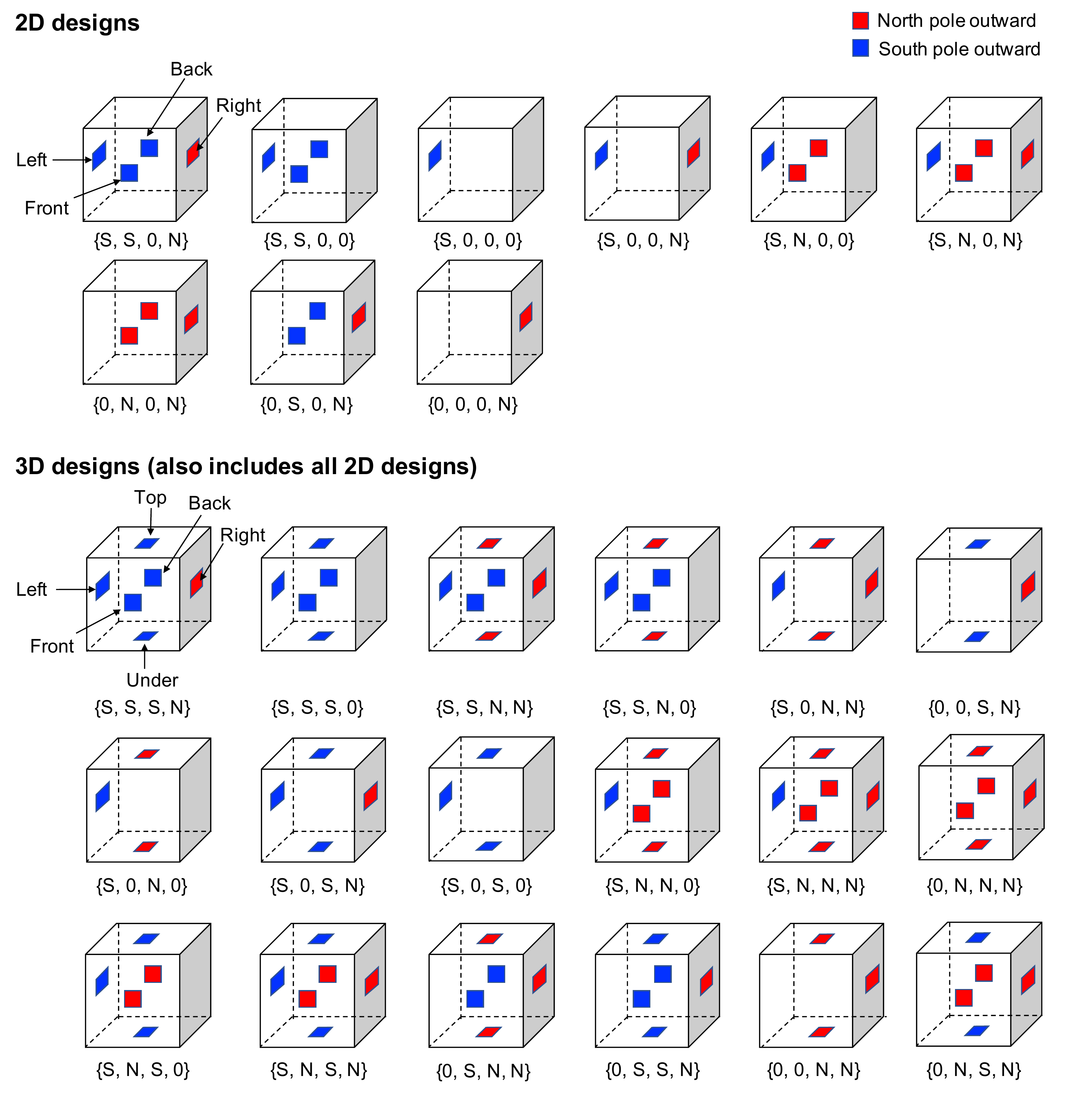}
\vspace{-1em}
\caption{Designs of 2D and 3D modular cubes.
Underneath each cube the arrangement is labeled as \{left [L], front [F]/back [B], top [T]/under [U], right [R]\}.\label{fig:CubesDesign}}
\vspace{-2em}
\end{figure}
 Much related work has focused on the ``Tilt model'' where components move in a straight line until they hit an obstacle, and combine when mating particles are brought close together. Solutions often design workspace obstacles that enable arbitrary reconfiguration of components pushed by a global force~\cite{balanza2019full,balanza2020hierarchical}.
 We designed algorithms that enabled efficient construction of desired shapes~\cite{becker2020tilt}, and workspace obstacles that enabled sorting and classifying constructed polyominoes~\cite{keldenich2018designing}. 

This previous work did not consider the stability or robustness of the structures generated. Recent work in~\cite{holobut2017distributed,piranda2020distributed} examined the forces between individual components and evaluated if a structure of module robots would be stable --- if it would topple or collapse under its own weight, especially during the process of reconfiguration. 

Our future goal is to design motion plans for magnetic modular cubes to enable them to form shapes suitable for fast motion, yet with designed fracture points so they can be easily disassembled at a goal and reconfigured into tools or structures. 
As first steps toward that goal, this paper presents the following contributions: 
(1) Design and demonstrations of magnetic modular cubes (MMCs) which are simple in design, scalable, and highly reconfigurable  through controlled assembly-disassembly. (2) Analysis of all possible magnetization profiles that match the specifications of our modular cubes. 
(3) A method to count the number of unique configurations for a given set of cubes. 
(4) Enumerate the unique configurations up to $n=20$ in 2D and $n=16$ in 3D.
(5) Provide a motion planner for 2D assembly.
(6) Monte Carlo simulations to measure the frequency that certain shapes can be constructed and the required number of movement steps to generate the shapes.
(7) Conduct hardware experiments that explore the self-assembly and disassembly of these modules in 2D and 3D. For applications of these magnetic cubes, see \cite{Bhattacharjee2021Mag}.

\section{Design of magnetic modular cubes}

We design the MMCs so they can be controlled by a global magnetic field. The cubes are manufactured so they align with an applied magnetic field.
The global magnetic field provides a reference frame with a magnetic south to the left and north to the right. We labeled each face of the cube as [L, F, B, T, U, R] for \emph{left}, \emph{front},  \emph{back}, \emph{top}, \emph{under}, and \emph{right} faces. On each face at most one axially magnetized permanent magnet is embedded. Each cube must have a net magnetic orientation. The magnets are positioned with poles facing outwards such that each face is either north or south and such that the net magnetic moment of the cube from magnetic south to north passes through the center of opposing faces.
This net magnetic moment is required because it enables ``pivot walking'' and ``rolling'' the cubes. Pivot walking is a movement gait that alternately lifts the north or south pole of the cube so the cube balances on the opposing edge, and then applies torque to spin the cube on the balancing edge, as in~\cite{Ehab,Rogowski2020}. The rolling  motion~\cite{bi2018design} can be achieved by applying continuous rotational magnetic torque along an axis perpendicular to the axis of net magnetization of a cube to roll it towards the desired direction. 

Because the magnetic field passes through the center of the cube face, the magnetic field must be rotationally symmetric across the left-to-right axis. 
Thus the magnetization of F==B and T==U. We can choose the L to be $(S$ or $0 )$, the F and B to be $(S,$ $0,$ or $N)$,  the T and U to be $(S,$ $0,$ or $N)$, and the R $(0$ or $N)$, for a total of $2\cdot3\cdot3\cdot2 = 36$ arrangements.
The arrangement is labeled as \{L, F/B, T/U, R\}.
We then remove the nine arrangements of the form $\{0,*,*,0\}$ because they have no net magnetic moment. Therefore, there are 27 sets of suitable magnet arrangements. 2D designs of MMCs only allow connection along the $x$-axis and $y$-axis. No magnets are embedded on the top [T] or under [U] faces. 
3D designs of MMCs enable connection along faces (see Fig.~\ref{fig:CubesDesign}).

Due to the MMCs’ cubic design, magnetically connected structures of MMCs are polyominoes in 2D and polycubes in 3D. Polyominoes and polycubes are a classic topic in combinatorics. This paper draws on a rich literature and results on enumerating polyominoes and polycubes~\cite{aleksandrowicz2006counting, luther2011counting, barequet2013polyominoes,  conway2016design, barequet2019improved} and on coloring them~\cite{demaine2019some}. Work on characterizing upper and lower bounds on the number of polyominoes and polycubes is an active research area~\cite{barequet2016lambda,barequet2019improved}, and will be relevant to modular robotic construction.

Similarly, there has been a great deal of recent research on how to design workspaces that exploit global control of magnetic particles to design arbitrary shapes~\cite{balanza2020hierarchical,caballerobuilding,cheang2016versatile}, and on the complexity of motion planning under such constraints. These papers provide important solutions on how to design workspaces that enable fast rearrangement of cubes, but they do not provide insight on how to build those cubes. Moreover, these works tend to assume all cubes will stick together if brought close enough in proximity. In contrast, this paper focuses on the configurations made possible by the magnetic profiles of the cubes.

\section{Generate all possible polyominoes in 2D}\label{sec:generateAllPolyominoes}
\renewcommand{\UrlFont}{\small\tt}

\emph{Polyominoes} are face-connected sets of unit cubes that lie on the square-grid graph. Each cube is represented by an integer tuple ($x$, $y$). Cubes ($x_1$, $y_1$) and ($x_2$, $y_2$) are adjacent if $|x_1-x_2|+|y_1-y_2|=1$. 
Two \emph{free polyominoes} are considered distinct if they have different shapes, but they cannot be rotations of each other. Two \emph{fixed polyominoes} are considered distinct if they have different shapes or orientations. Because the global magnetic field provides an orientation, we consider fixed polyominoes for our MMCs design. The number of fixed polyominoes of size $n$ is denoted by \AtwoD.  The number of polyominoes grows exponentially~\cite{barequet2016lambda} 
\begin{equation}
\label{eq:polyominoGrowRate}
\lim_{n\to\infty}(\AtwoD)^{\frac{1}{n}} = \lambda.
\end{equation}

\begin{figure}[t]
\centering
\includegraphics[width=1\linewidth]{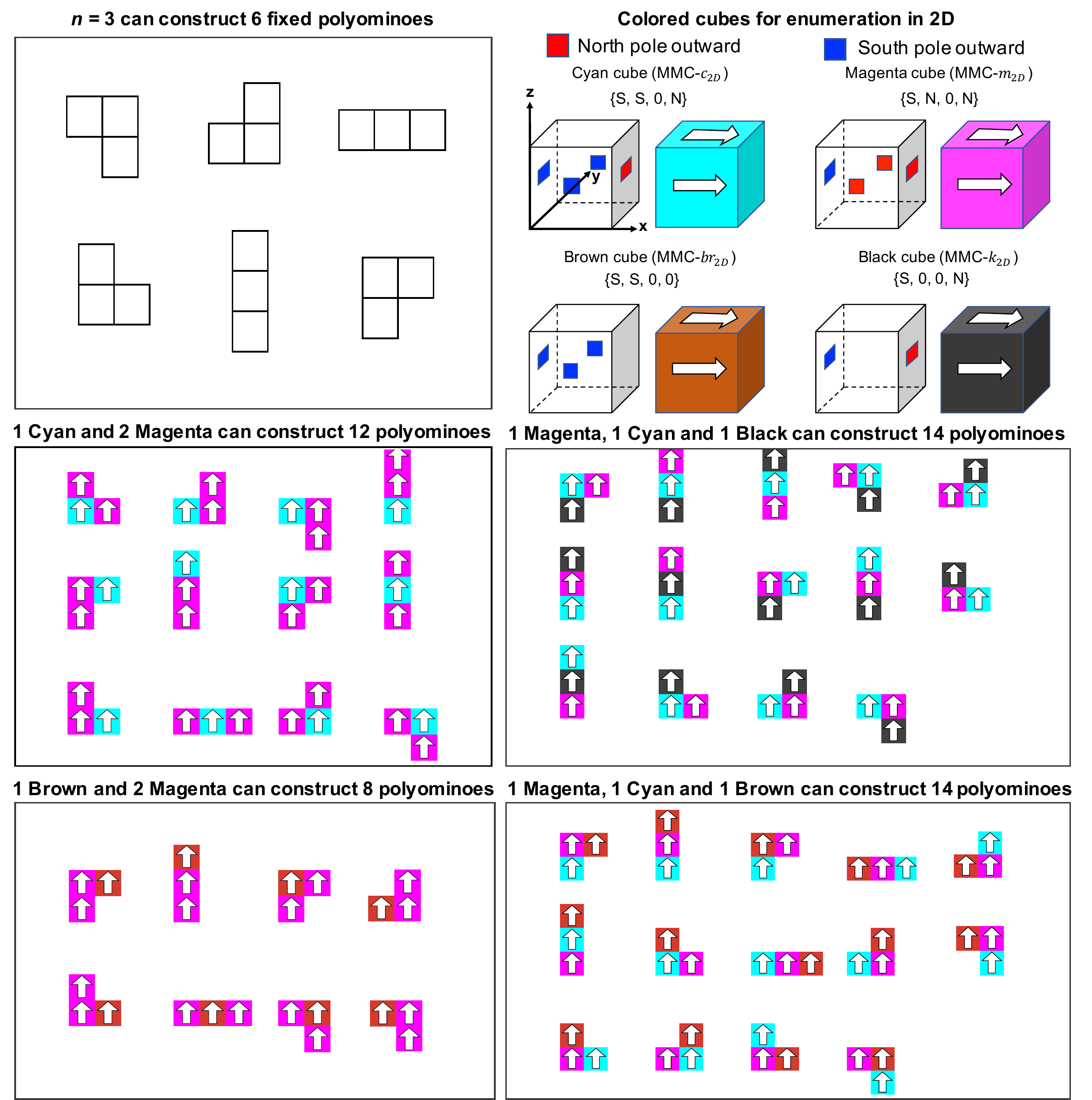}
\vspace{-1em}
\caption{Enumeration of colored cubes in 2D. The composition of cube types determines the number of polyominoes that can be made. The white arrow indicates the direction of south to north magnetic pole.\label{fig:2D colored cube}}
\vspace{-1.5em}
\end{figure}

\begin{table*}[t]
\centering
\caption{Number of fixed polyominoes \AtwoD, where ${n}$ is the number of cubes. With MMC-m$\textsubscript{2D}$ and MMC-c$\textsubscript{2D}$, ${C_{n,i}}$ is the number of colored polyominoes with ${i}$ magenta cubes and $n$ total cubes. Row maximum in \textcolor{myorange}{orange}.\label{fig:2D}
}
\includegraphics[width=0.95\linewidth]{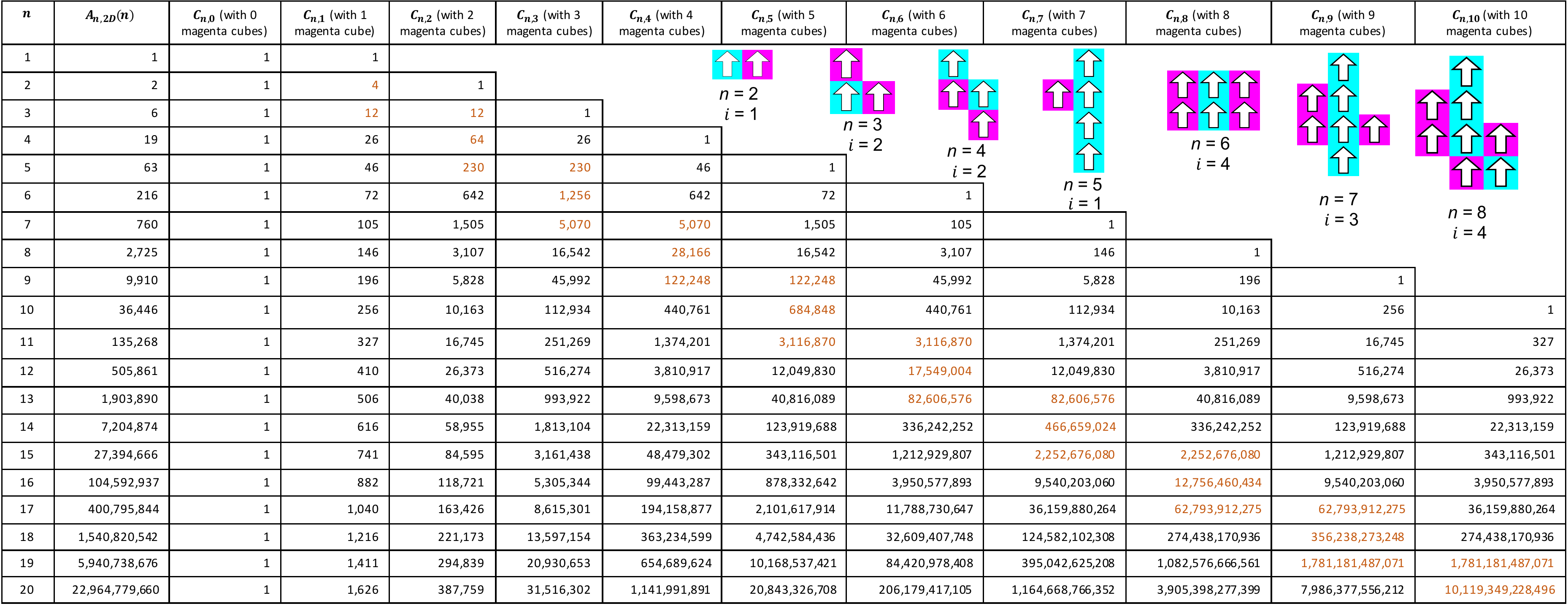}
\vspace{-2em}
\end{table*}

We implemented a method to enumerate the fixed polyominoes and unique colored polyominoes that can be constructed by a set of MMCs~\cite{yitongEnumerateCode}. Based on the hardware experiment environment (see Section~\ref{sec:experimentalResults}), we developed a 2D motion planner using a bounded 11$\times$11 workspace that generates the set of valid colored polyominoes reachable by global moves from an initial configuration. The cubes are represented by axis-aligned, unit-length tiles in our motion planner. Arrows on the tiles indicate the direction of the net magnetization (from magnetic south pole to north pole). We select four magnet arrangements that use at least two magnets and, for ease of visualization, assign them the colors cyan, magenta, brown, and black (see Fig.~\ref{fig:2D colored cube} top right).

We enumerate the fixed polyominoes from a given number of cubes, then calculate the set of valid colored polyominoes that can be generated from a given set of MMCs. We determine the shortest movement sequences to generate all reachable polyominoes from a given initial configuration. For each polyomino, we analyze the fraction of starting configurations for which it is reachable and the number of moves required to construct it.

\subsection{Enumeration of fixed polyominoes}
To enumerate fixed polyominoes, we implemented the method used by Redelmeier in~\cite{redelmeier1981counting}. 
We use a variant of this algorithm to enumerate colored polyominoes constructed from MMCs. Because the magnets in these cubes define a coordinate system, we enumerate all polyominoes with magnetic moments pointing in the $+y$ direction. 
The algorithm works by recursively generating all cubes up to a given size $n$. We start with the leftmost cube of the bottom row, placed at the origin $(0,0)$. For a given configuration with $k$ cubes, the algorithm generates a new configuration of size $k+1$ by adding a cube connected to the polyomino. We maintain a list of the adjacent cubes, \emph{list\_adj}, and a list of which adjacent cubes have been selected, \emph{poly}, for each recursive call. In the initial call, the list of adjacent cubes, \emph{list\_adj}, consists of the origin $(0,0)$, the cube above the origin $(0,1)$, and the cube to the right of the origin $(1,0)$. We set \emph{poly}[0] = 0. At each recursive call, if $k\equiv n$, the procedure returns. Otherwise, a recursive call is made for each cube in \emph{list\_adj} greater than the last cube in \emph{poly}. For each of these calls, the cube is added to \emph{poly}, and any cubes adjacent to the new cube are appended to \emph{list\_adj} if they are not already in \emph{list\_adj}. To avoid generating the same cube more than once, cubes are only added if they are above the origin, or in the same row to the right of the origin: 
\begin{equation}
\label{eq:add adj}
\left\{ (x,y)~|~(y>0) \textrm{ or } (y\equiv 0 \textrm{ and } x\geq0) \right\}
\end{equation}

\subsection{Enumeration of valid colored polyominoes from a set}
To count the number of valid colored polyominoes from a set of colored cubes, we modify the previous algorithm. We include the coloring information in \emph{poly}. Before each recursive call, we check if the new cube can be colored in the given color. A check is successful if there are sufficient unused cubes of the color, and if placing the cube does not violate any of the magnetic-assembly rules. If the check is successful, a recursive call is performed. Fig.~\ref{fig:2D colored cube} top left shows that for $n=3$ cubes, 6 fixed polyominoes can be enumerated. Using a supply with one cyan and two magenta cubes produces more polyominoes than one brown and two magenta cubes. However, using three colored cubes enables constructing even more colored polyominoes.

\begin{figure*}[t]
\centering
\includegraphics[width=0.95\linewidth]{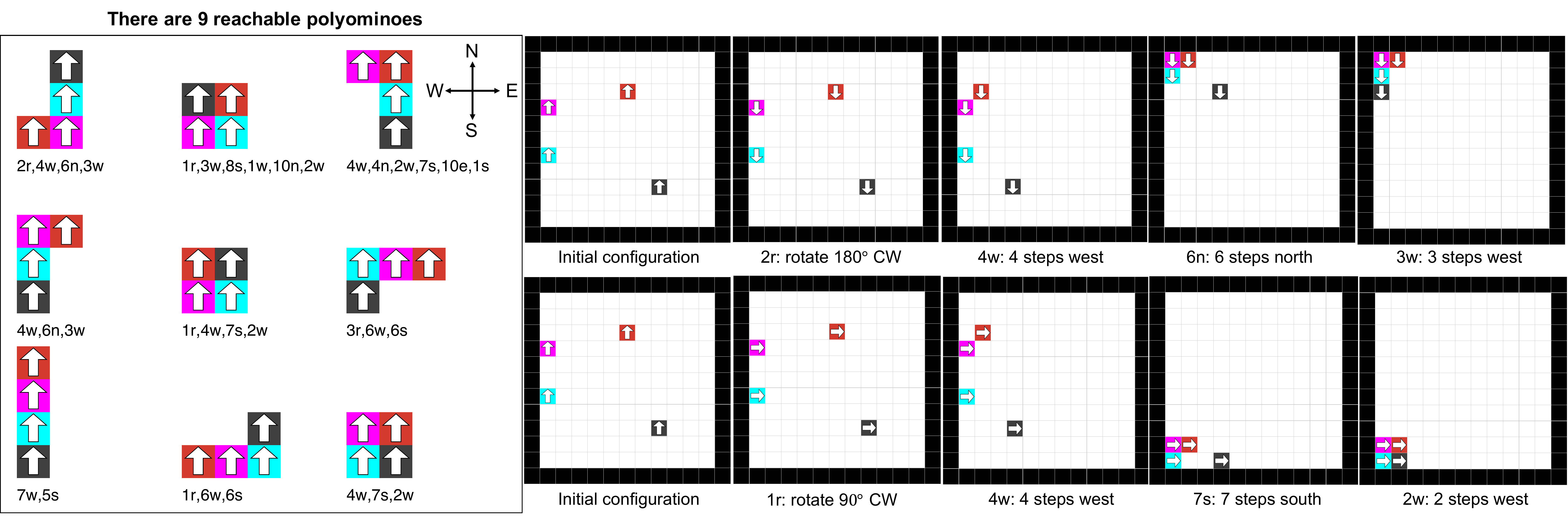}
\caption{\textbf{Left}: The self-assembly result for an arbitrary initial configuration of four cubes with different designs. \textbf{Right}: Details of the movements.\label{fig:polyominoes}}
\end{figure*}

\begin{figure*}[t]
\centering
\includegraphics[width=1\linewidth]{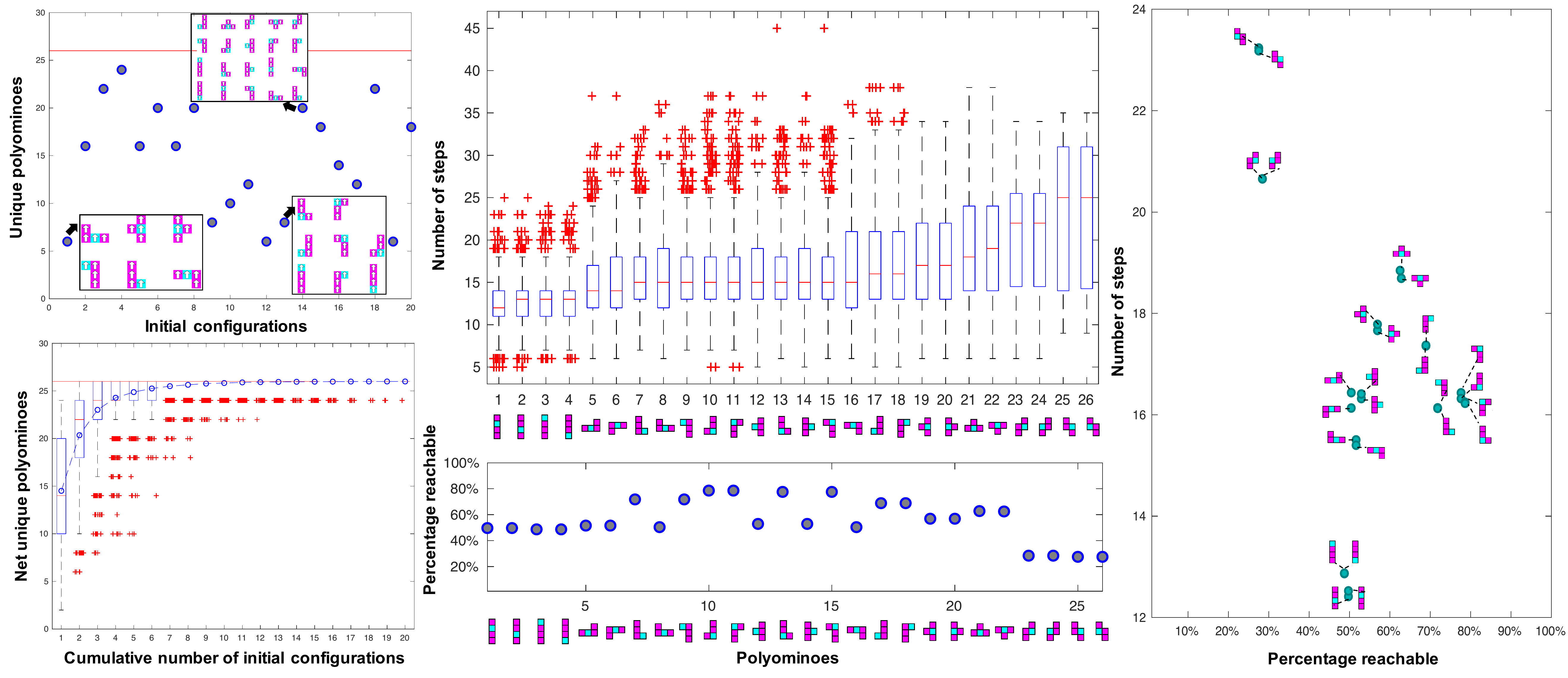}
\caption{\textbf{Top Left}: The number of polyominoes that can be made with 1 cyan and 3 magenta cubes from random starts. The red line is the maximum number of possible configurations (26).
This test was repeated for 1000 trials, and analyzed in the remaining plots. 
\textbf{Bottom Left}: Cumulative number of unique reachable polyominoes. 
Red + signs show outliers. \textbf{Middle Top}: Total number of steps to construct each polyomino (when the polyomino is reachable). \textbf{Middle Bottom}: Frequency results showing how often each polyomino is reachable. \textbf{Right}: Scatter plot of percentage reachable vs. number of steps.\label{fig:3r1bpolyominoes}
}
\vspace{-1.5em}
\end{figure*}

Table~\ref{fig:2D} shows the number of fixed polyominoes that can be enumerated with up to $n=20$ cubes. It also lists the number of valid polyominoes with magenta and cyan cubes that can be constructed by coloring them with up to 20 cyan cubes. 
With zero to two magenta cubes, the number of valid polyominoes is less than the number of fixed polyominoes ($C_{n,i}$ for $i$ = 0, 1 or 2). This is because two magenta or two cyan cubes can only be connected in series (with arrows aligned tip-to-tail). However, a more extensive set of polyominoes with different shapes and orientations can be constructed when the fraction of magenta cubes is increased. The largest variety is possible when $i$ = $\lfloor \frac{n}{2} \rfloor$. The top right corner of Table~\ref{fig:2D} shows examples of valid colored polyominoes.

\subsection{Self-assembly algorithm}
We provide a low-fidelity motion model that can compute reachable polyomino configurations and the shortest movement sequences for a set of MMCs from an initial configuration. We assume that all modules move at the same speed in the same direction, unless they encounter a fixed obstacle. Without loss of generality, we limit the movement directions to north, east, south, and west.

Fig.~\ref{fig:polyominoes} shows an example of all the valid colored polyominoes can be generated with 4 different cube designs. The initial configuration and the details of the movements are shown on the right. Underneath each polyomino is listed the shortest step sequence to construct it. At each step, we could move one unit length in the four cardinal directions \{n, e, s, w\} or rotate \{r\} counterclockwise (CCW) or clockwise (CW) a quarter-turn. We prune this search by only allowing rotations if no translation moves have been made and by only performing CW rotations. For example, in this shorthand, ``2r, 4w, 6n” means rotate 180$^\circ$ CW followed by four unit moves west, and then six unit moves north. We used a breadth-first search (BFS) algorithm to discover these reachable configurations. 
The root of the search is the initial configuration, and we maintain a list of all relative configurations that have been reached. A \emph{relative configuration} is represented by rotating the coordinate frame so that the magnetic north points up and then translating all cubes. To further prune the tree, we only switch direction if the relative configuration of the cubes changes, and we terminate search branches if the present configuration has occurred before. When the relative configuration changes, we call this an \emph{intermediate configuration}. Intermediate configurations occur when at least one cube (but not all cubes) strikes an obstacle. When this occurs, the new relative configuration is compared to the list and is only appended if it is unique. After the polyomino is assembled, we compare the paths taken to save the shortest step sequence to construct the structure.

\begin{table*}[t]
\centering
\caption{Number of fixed polycubes \AthreeD, where ${n}$ is the number of cubes. Row maximum in \textcolor{myorange}{orange}.\newline With MMC-r$\textsubscript{3D}$ and MMC-b$\textsubscript{3D}$, ${C_{n,i}}$ is the number of colored polycubes with ${i}$ red cubes and $n$ total cubes.\label{tab:3D Blue and Red}}
\includegraphics[width=1\linewidth]{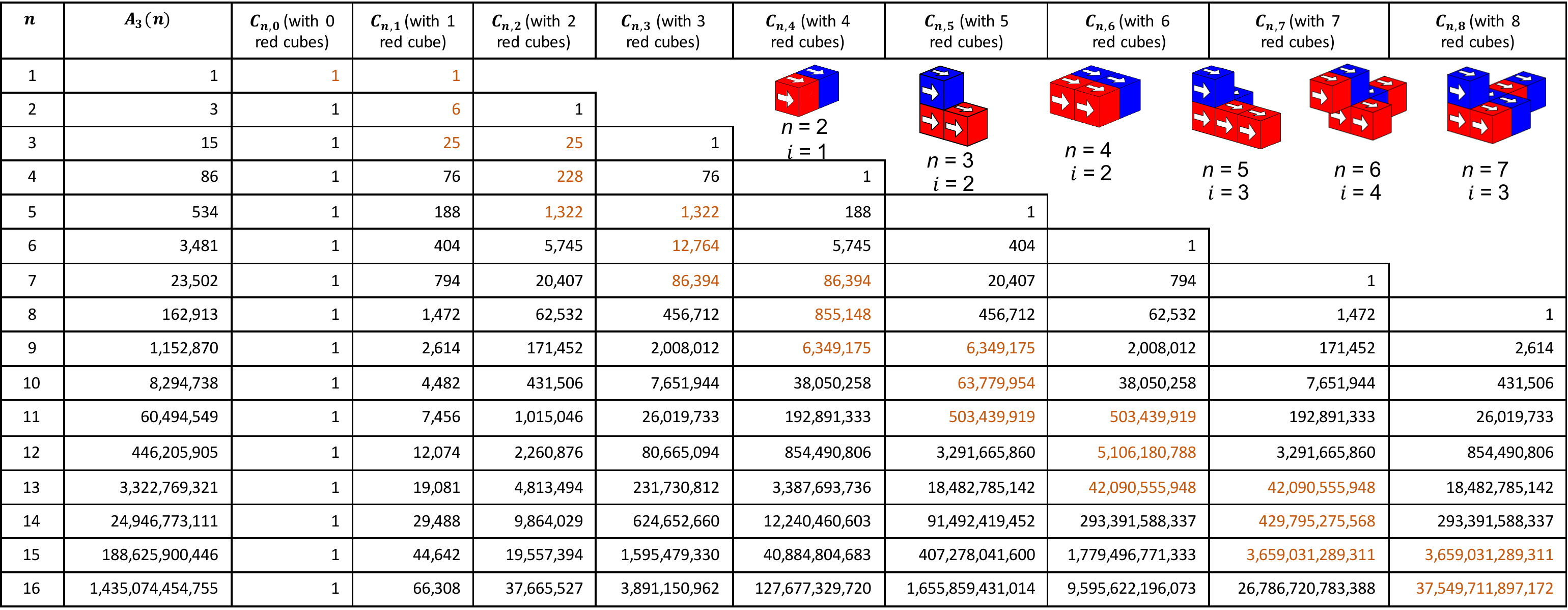}
\vspace{-2em}
\end{table*}

The number of reachable polyominoes depends on the initial configuration.
For a given number of magenta and cyan cubes, a finite number of  colored polyominoes can be generated (see Table~\ref{fig:2D}), but some may be unreachable from an arbitrary initial configuration. Different initial configurations usually generate different numbers of reachable polyominoes. Fig.~\ref{fig:3r1bpolyominoes} top left shows examples of 20 random initial configurations. For each configuration, the number of unique reachable polyominoes was calculated. Three black boxes show the valid reachable polyominoes that can be generated from three initial configurations. However, from none of these 20 configurations could we construct every one of the 26 valid colored polyominoes (see Table~\ref{fig:2D}). If the desired polyomino is unreachable from a given initial configuration, one could repeatedly generate new random initial configurations until the polyomino is reachable. Rapidly rotating the magnetic field is one way to separate and scramble the cubes into new configurations.

To test the feasibility of this, we ran 1000 trials using one cyan and three magenta cubes. Fig.~\ref{fig:3r1bpolyominoes} bottom left shows the net unique polyominoes that can be constructed after a given number of configurations. The median number of reachable polyominoes from the first start is 14. With two restarts, it is possible to cover on average 20.28 of the 26 possible polyominoes configurations. In 999 out of 1000 trials, all 26 polyominoes were reachable after 20 accumulated configurations. The middle top figure shows the number of steps needed to construct each polyomino (if the polyomino is reachable) over 1000 trials. For these trials, we only counted translation steps and did not count rotation steps because translations require multiple changes to the magnetic field. The middle bottom figure shows the percentage reachable results for each polyomino. Icons at the bottom of these figures show each valid polyomino. I-shapes require on average the fewest steps to form. L-shapes with a cyan cube in the corner are reachable most often (the highest frequency is 786 out of 1000 trials). Z-shapes are harder to make because they require assembly in a specific temporal order (moving all cubes to give them the same $y$ coordinate, so to make a Z-shape, the motion must be stopped before all cubes reach the boundary). Z-shapes also require on average the most steps to form (see Fig.~\ref{fig:3r1bpolyominoes} right).

\section{Generate all possible polycubes in 3D}
A \emph{polycube} is a three-dimensional polyomino constructed by cubes attaching face to face. Each cube is represented by an integer tuple $(x, y, z)$. Cubes 1 and 2 are adjacent if $|x_1-x_2|+|y_1-y_2|+|z_1-z_2|=1$. Polycubes can be enumerated in two ways, depending on whether different orientations are counted as one polycube or two. Two \emph{free polycubes} are considered distinct if they have different shapes, but are not 3D rotations of each other. Two \emph{fixed polycubes} are considered equivalent if one can be transformed into the other by a translation. The number of fixed three-dimensional polycubes of size $n$ is denoted by \AthreeD.

\begin{figure}[t]
\centering
\includegraphics[width=1\linewidth]{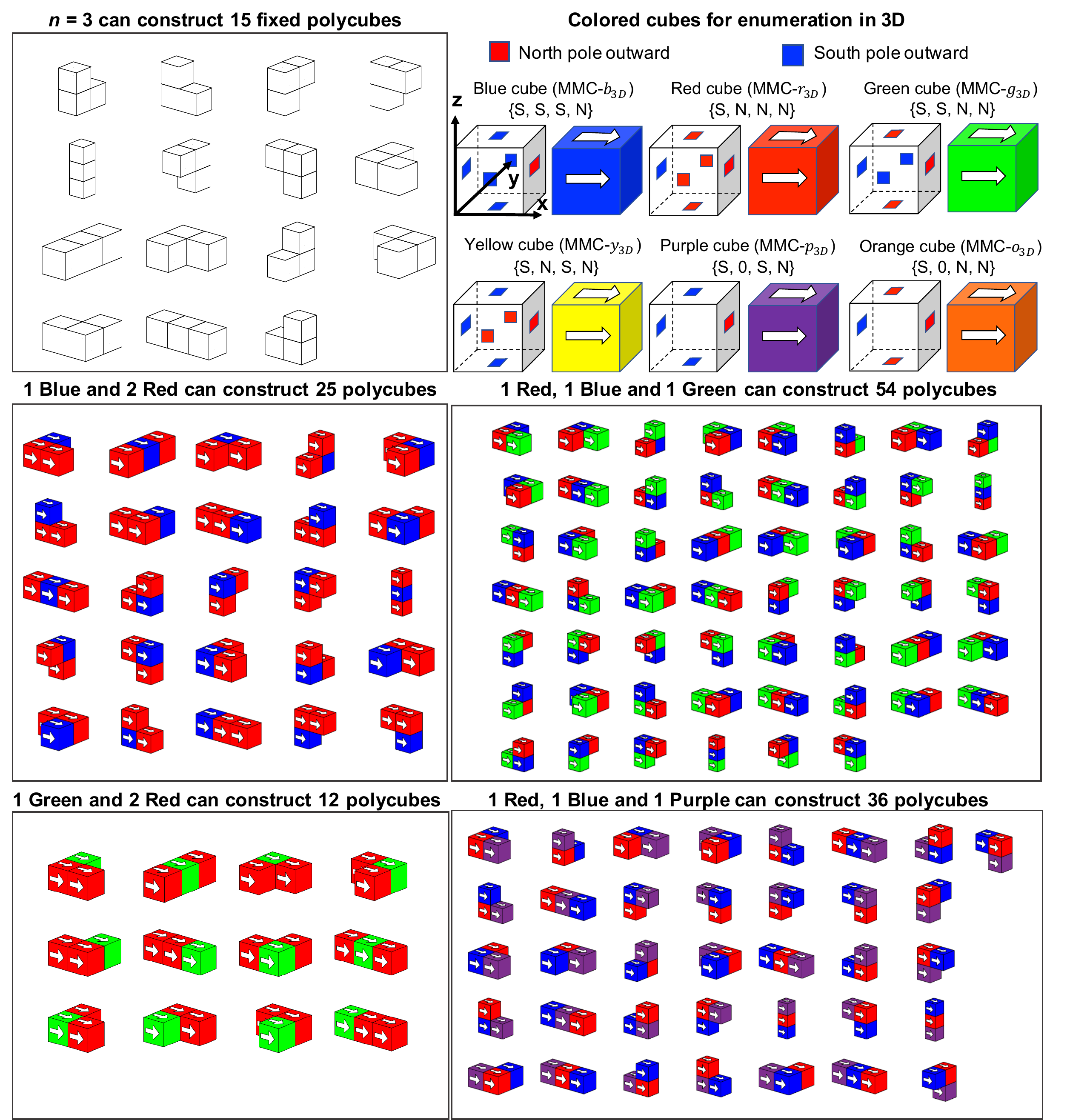}
\caption{Enumeration of colored cubes in 3D. The composition of cube types determines the number of polycubes that can be made.  
The white arrow indicates the direction of south to north magnetic pole. \label{fig:3D colored cube}}
\vspace{-2.2em}
\end{figure}

\begin{figure*}[t]
\centering
\includegraphics[width=1\linewidth]{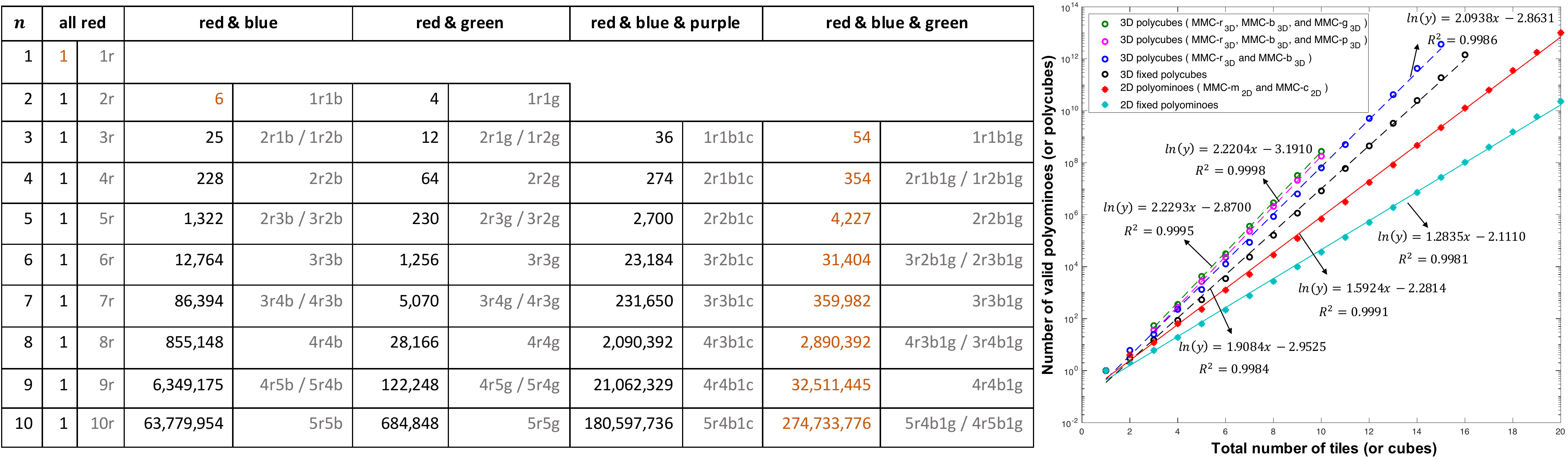}
\caption{\textbf{Left}: Maximum number of valid colored 3D polycubes with different sets of cube colors, where ${n}$ is the total number of cubes. \textbf{Right}: Enumeration results of valid colored polyominoes and polycubes.\label{tab:3Dmulticolors}}
\vspace{-2em}
\end{figure*}

\subsection{Enumeration of fixed polycubes}
As with the enumeration of polyominoes in 2D, we start with the leftmost cube of the bottom row and place it at the origin $(0,0,0)$. 
In the initial call, the list of adjacent cubes, \emph{list\_adj}, consists of the origin $(0,0,0)$, the cube to the right of the origin $(1,0,0)$, the cube to the back of the origin $(0,1,0)$, and the cube on the top of the origin $(0,0,1)$, we set \emph{poly}[0] = 0. 
To avoid double-counting polycubes, cubes are only added at: 
\begin{align} \label{eq:3Daddadj}
\big\{ (x,y,z) ~|~ &(z>0) \textrm{ or } (z\equiv 0 \textrm{ and } y>0) \nonumber \\
&\textrm{ or } (z\equiv 0 \textrm{ and } y\equiv 0 \textrm{ and } x>0)  \big\}.
\end{align}

\subsection{Enumeration of valid colored polycubes from a set}\label{subsec:coloring}
We select six magnet arrangements that use at least four magnets and assign them the colors in Fig.~\ref{fig:3D colored cube} top right. As with the enumeration of valid colored polyominoes in 2D, to count the number of valid colored polycubes generated from a set of MMCs, we color the cubes while enumerating fixed polycubes during the recursive step. Fig.~\ref{fig:3D colored cube} top left shows that for $n=3$ cubes, 15 fixed polycubes can be enumerated. Middle left and bottom left figures show examples of coloring with two colors. Using a supply with red and blue cubes can produce more polycubes than red and green. With a supply of three colors, a wider variety of valid colored polycubes can be constructed.

Table~\ref{tab:3D Blue and Red} shows the number of fixed polycubes that can be enumerated with up to $n=16$ cubes and the number of valid polycubes that can be constructed by coloring them with red and blue cubes. The top right corner of the table shows examples of valid colored polycubes.
Fig.~\ref{tab:3Dmulticolors} left shows the maximum number of valid polycubes that can be constructed with five different sets of cubes in 3D, where ${n}$ is the total number of cubes. For each cube set, the left column shows the maximum number of valid colored polycubes that can be generated for a given total number of cubes. The right column in each set (gray font) shows configurations in a shorthand, e.g. ``1r4b" means 1 red and 4 blue cubes.

\subsection{Rejecting non-magnetically connected components}\label{sec:connected-component}
\begin{figure}[t]
\centering
\includegraphics[width=0.95\linewidth]{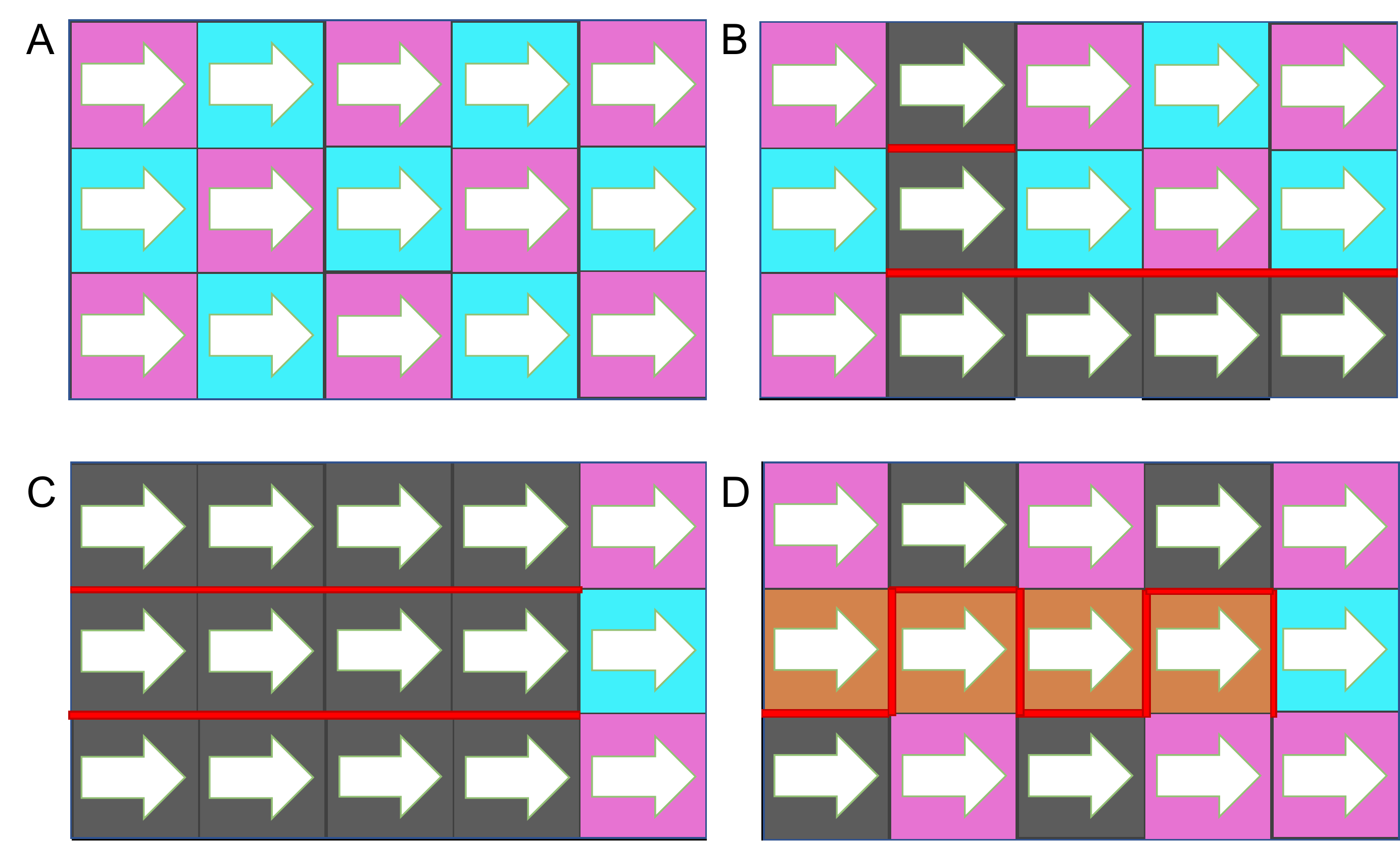}
\caption{Examples of magnetically connected polyominoes with the same shape but different bonds. Red lines show potential failure surfaces with no magnetic connection.\label{fig:connection}}
\vspace{-2em}
\end{figure}

Our enumeration of valid polyominoes (or polycubes) allows placing a cube  face without magnets next to a cube of any color, but there is no magnetic connection. Our goal for the polyominoes (or polycubes) is to create aggregates that can be manipulated in unison by a magnetic field. For the purpose of this study, we require that every cube be magnetically connected to form a single component. This requires connected-component analysis. Connected-component labelling is normally used for distinguishing different objects in a binary image~\cite{he2017connected}. A \emph{magnetically-connected component} is a set of cubes, $C$, such that for any two cubes in $C$, there is a valid connected path between them and this path is contained in $C$. A \emph{valid connection} between two MMCs means the two cubes share a face and on this face one cube has a south pole outward and the other has a north pole outward. Before adding a polycube of size $n$ to our enumeration, we first check that the polycube is a single magnetically-connected component.
We start with the $(0,0,0)$ cube, and perform a breadth-first search on all valid-connected neighboring cubes. If the total number of connected cubes is $n$, the polycube is counted, otherwise it is rejected. All the polyominoes in Fig.~\ref{fig:connection} are the same shape and are magnetically connected, but the lack of magnetic bonds introduces planar cracks in the assembly, which could be exploited for controlled disassembly. Future work could catalog these shapes during the enumeration process.

\subsection{Time complexity}
As with the enumeration scheme by Redelmeier, our enumeration counts each possible polyomino and polycube only one time, the enumeration complexity is exponential, $O(c_{\text{2D}}^n)$ and $O(c_{\text{3D}}^n)$. Our implementation is slightly less efficient if we allow cubes that do not have magnets on each side, since we generate intermediate configurations that might not form a magnetically-connected component. After constructions, the final step is to check if the polycube is magnetically-connected, and reject those that are not. The connected check requires $O(n)$ time. 3D construction is more complex than 2D, so the constant $c_{\text{3D}} > c_{\text{2D}}$. Fig.~\ref{tab:3Dmulticolors} right shows the results of enumeration of valid colored polyominoes and polycubes.

\section{Experimental Setup and Results} \label{sec:experimentalResults}
A large-scale nested triaxial Helmholtz coil system was used to conduct  experiments. This system was designed parametrically to create uniform magnetic fields in $x$, $y$, and $z$. Each coil pair is connected to its own programmable power supply
which is controlled by a National Instruments data acquisition (DAQ) board. The power supplies generate sinusoidal outputs to the coils, which in turn creates a uniform rotating magnetic field with user-specified magnitude and frequency. These signals are  controlled using a customized C++ program. The 3D magnetic field vector produced from the system is represented by
\begin{equation}
\textbf{\textit{B}} =
    \begin{bmatrix}
    B_x\\
    B_y\\
    B_z\\
    \end{bmatrix} 
    = \begin{bmatrix}
  A\cos(\alpha)\cos(\theta)\\
    A\cos(\alpha)\sin(\theta)\\
    A\sin(\alpha)\\
    \end{bmatrix},\label{eq:Magnetic Flux Density}
\end{equation} where \textbf{\textit{B}} is the applied magnetic flux density, $B_x$, $B_y$, and $B_z$ are the three-dimensional components along $x$, $y$ and $z$-axes respectively, $A$ is the amplitude, $\alpha$ is the pitch angle, and $\theta$ is the yaw angle. Individual cubes that have a volume of 1cm$^3$ were designed using a CAD
software (Onshape) and printed using polylactic acid (PLA) on a 3D printer (Ultimaker 2 Extended+). A {\sc Matlab} program collected magnetic field settings throughout the experiment and recorded assembly behavior of modular subunits with a digital camera.

\begin{figure}[t]
\centering
\includegraphics[width=0.9\linewidth]{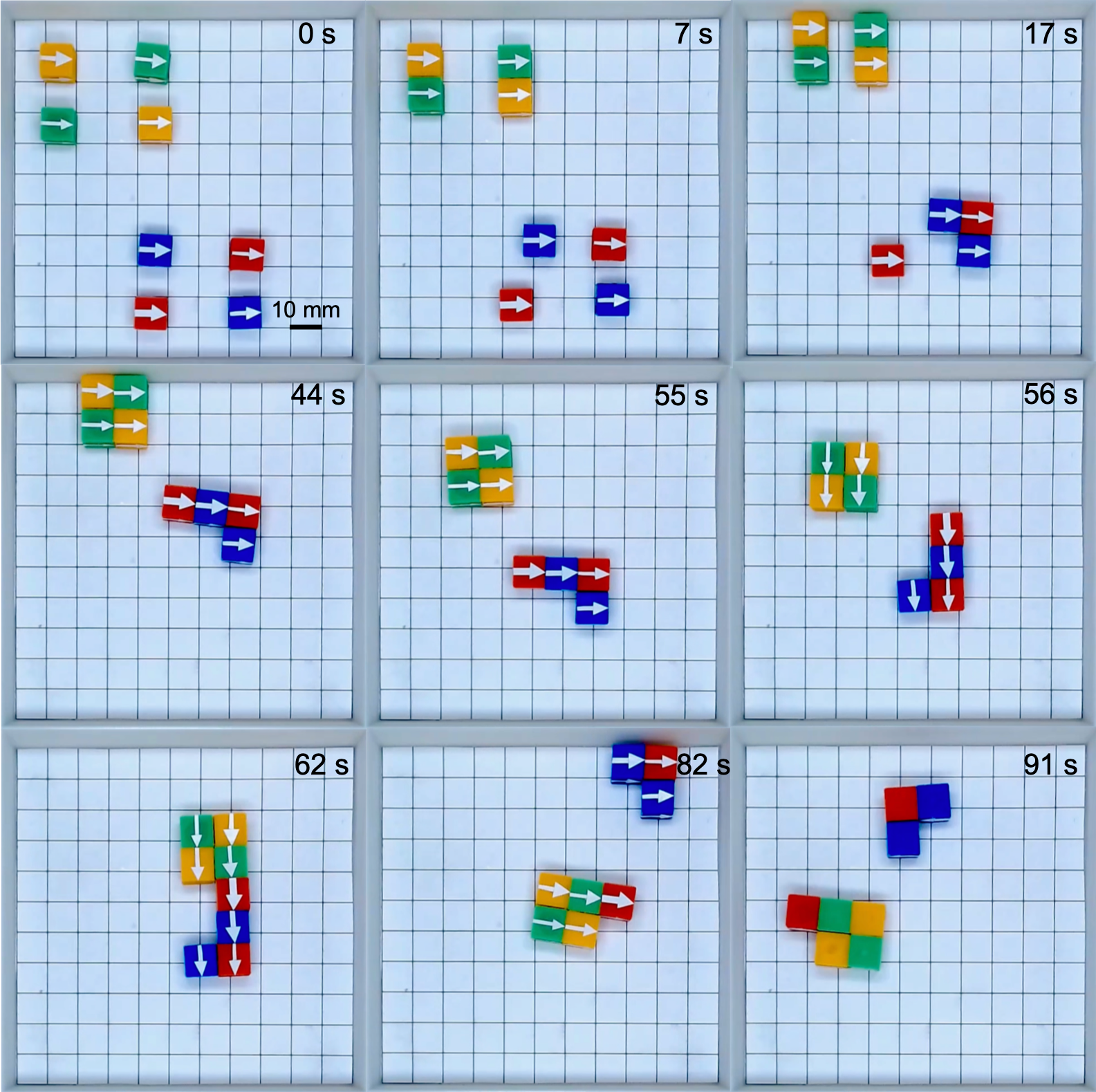}
\caption{2D self-assembly and disassembly behavior of modular subunits with eight cubes.\label{fig:2Dexperiment}}
\vspace{-1.5em}
\end{figure}

A magnetic flux of 10mT was enough to actuate the modular cubes and achieve successful assembly. Pivot walking motion was used to actuate individual modular cubes with precise motion, following predetermined paths to create a target assembly. When the arrows on top faces of the modular cubes are aligned side-by-side, this is defined as a parallel assembly. When the arrows are aligned tip-to-tail, this is defined as a serial assembly. The self-assembly is caused by the magnetic attraction force generated from embedded permanent magnets in the modular cubes when they are moved close to each other. Fig.~\ref{fig:2Dexperiment} shows 2D self-assembly and disassembly behavior of modular subunits for eight cubes. Starting with two groups of modular subunits, with an equal number of modular cubes in each group, serial and parallel assembly were explored to form square- and \reflectbox{L}-shapes (at 44s). However, the self-assembly configuration depends on the motion path taken (i.e., parallel assembly or serial assembly) and the choice of subunits. From 44s to 55s, the assembled structures could be moved downwards by pivot walking, like an individual modular cube. From 55s to 56s,  the sub-assemblies were rotated  $90^\circ$ CW by changing the direction of the static magnetic field. The sub-assemblies could then be moved right or left to bring them into close proximity. The sub-assemblies were then joined to create a large 9-shaped structure (at 62s). Then, it was disassembled into two different shapes (at 82s), with a rapid $90^\circ$ CCW rotational magnetic torque. The disassembly occurred at the weakest  magnetic joint in the structure.  The modular structures can also be moved in the workspace using the rolling motion~\cite{bi2018design} which allows flipping the structures (at 91s).

\begin{figure}[t]
\centering
\includegraphics[width=0.9\linewidth]{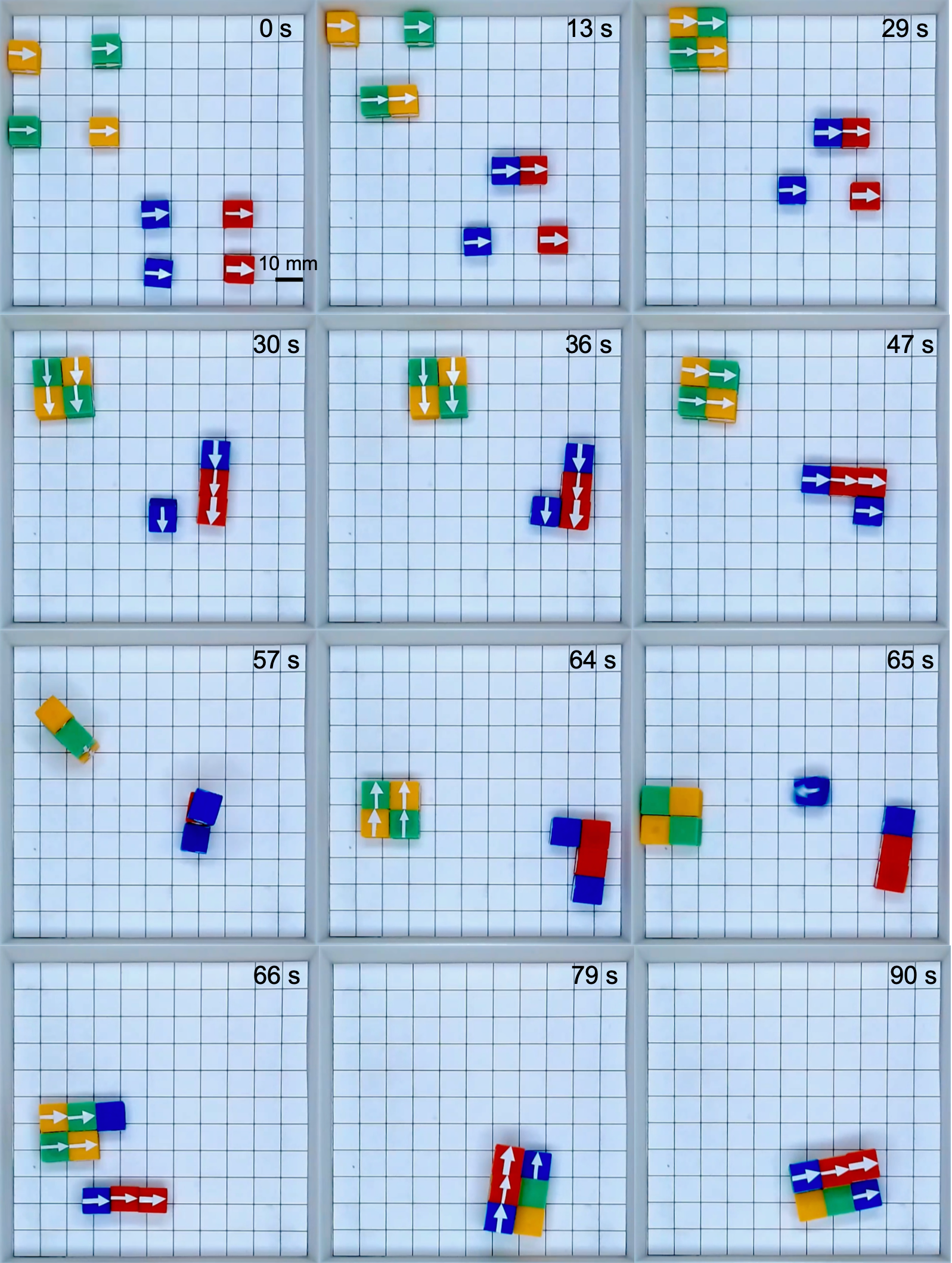}
\caption{3D self-assembly and disassembly behavior of modular subunits with eight cubes.\label{fig:3Dexperiment}}
\vspace{-2em}
\end{figure}

Fig.~\ref{fig:3Dexperiment} shows a representative 3D construction sequence using global magnetic control. A 3D self-assembly was performed following a 2D self-assembly to create a pair of sub-assemblies. Starting with a similar configuration as Fig.~\ref{fig:2Dexperiment}, but choosing a different motion path in the 2D self-assembly process formed an \reflectbox{L}-shape and a square-shape with different cube orderings at 36s. The sub-assemblies were rotated CW (at 30s) and CCW (at 47s) for movement in different directions. Once 2D structures were formed with  pivot walking motion followed by 2D self-assembly, the sub-units can be further joined in 3D by using rolling motions. The rolling motion was achieved by applying continuous rotational magnetic torque on the modular structures about a fixed axis until the target location was reached (from 57s to 66s). While the sub-assemblies were rolled to bring them in close proximity, a blue modular cube in the \reflectbox{L}-shaped structure was disassembled (at 65s). The blue cube reattached to the square-shaped structure (at 66s) which resulted in an unplanned reconfiguration. This spontaneous disassembly and reconfiguration could have benefits, but could also be avoided with improved control. Once the reconfigured sub-units move into close proximity, they self-assembled to create a two layered 3D structure (at 79s). This 3D structure was also actuated in the workspace with the pivot walking motion (from 79s to 90s).

\section{Conclusion}
This paper describes the 2D and 3D configurations of modular cubes with magnetic faces controllable by a global magnetic field. Experimental results demonstrate the self-assembly and disassembly behavior of magnetic modular cubes in 2D and 3D. Computational modeling enumerates the set of possible polyominoes and polycubes from a set of modular cubes. The 2D motion planner computes the shortest movement sequences to generate all reachable polyomino configurations. 

Future work will focus on 3D planning, in particular making a high-fidelity simulator for motion planning. It would be interesting to calculate the number of valid colorings of a given polycube as a function of the supply of colored cubes.  During the enumeration process one could also catalog useful configurations, ranking them on their mechanical stability, fracture modes, or magnetic profile in addition to their shape.

\bibliography{biblio}
\bibliographystyle{IEEEtran}

\end{document}